\pdfoutput=1

\documentclass[11pt]{article}

\usepackage[final]{acl}

\usepackage{times}
\usepackage{latexsym}

\usepackage[T1]{fontenc}

\usepackage[utf8]{inputenc}

\usepackage{microtype}

\usepackage{inconsolata}

\usepackage{graphicx}
\usepackage{amsmath}
\usepackage{algorithm}
\usepackage{algorithmic}
\usepackage{latexsym}

\usepackage{microtype}

\usepackage{booktabs}
\usepackage{multirow}
\usepackage{mathtools}
\usepackage{subfigure}
\usepackage{balance}
\usepackage{color}
\usepackage{xcolor}
\usepackage{enumitem}
\usepackage{xspace}
\usepackage{setspace}
\usepackage{amssymb}
\usepackage{longtable}

\usepackage{xcolor}     
\usepackage[markup=underlined]{changes}

\title{Adaptive Schema-aware Event Extraction with Retrieval-Augmented Generation}

\author{%
  Sheng Liang\textsuperscript{1}, Hang Lv\textsuperscript{2}, Zhihao Wen\textsuperscript{1}, Yaxiong Wu\textsuperscript{1}, Yongyue Zhang\textsuperscript{1},\\ \textbf{Hao Wang\textsuperscript{2}, Yong Liu\textsuperscript{1}} \\
  \textsuperscript{1}Huawei Noah’s Ark Lab, \textsuperscript{2}University of Science and Technology of China
\\
  \texttt{liangsheng16@huawei.com} \\
}

\begin{document}
\maketitle
\begin{abstract}
Event extraction (EE) is a fundamental task in natural language processing (NLP) that involves identifying and extracting event information from unstructured text. Effective EE in real-world scenarios requires two key steps: selecting appropriate schemas from hundreds of candidates and executing the extraction process.
Existing research exhibits two critical gaps: 
(1) the rigid schema fixation in existing pipeline systems, and (2) the absence of benchmarks for evaluating joint schema matching and extraction.
Although large language models (LLMs) offer potential solutions, their schema hallucination tendencies and context window limitations pose challenges for practical deployment. 
In response, we propose \textbf{A}daptive \textbf{S}chema-aware \textbf{E}vent \textbf{E}xtraction (\textbf{ASEE}), a novel paradigm combining schema paraphrasing with schema retrieval-augmented generation. ASEE adeptly retrieves paraphrased schemas and accurately generates targeted structures.
To facilitate rigorous evaluation, we construct the \textbf{M}ulti-\textbf{D}imensional \textbf{S}chema-aware \textbf{E}vent \textbf{E}xtraction (\textbf{MD-SEE}) benchmark, which systematically consolidates 12 datasets across diverse domains, complexity levels, and language settings.
Extensive evaluations on MD-SEE show that our proposed ASEE demonstrates strong adaptability across various scenarios, significantly improving the accuracy of event extraction.

\end{abstract}

\section{Introduction}

\begin{figure}[t]
  \includegraphics[width=\columnwidth]{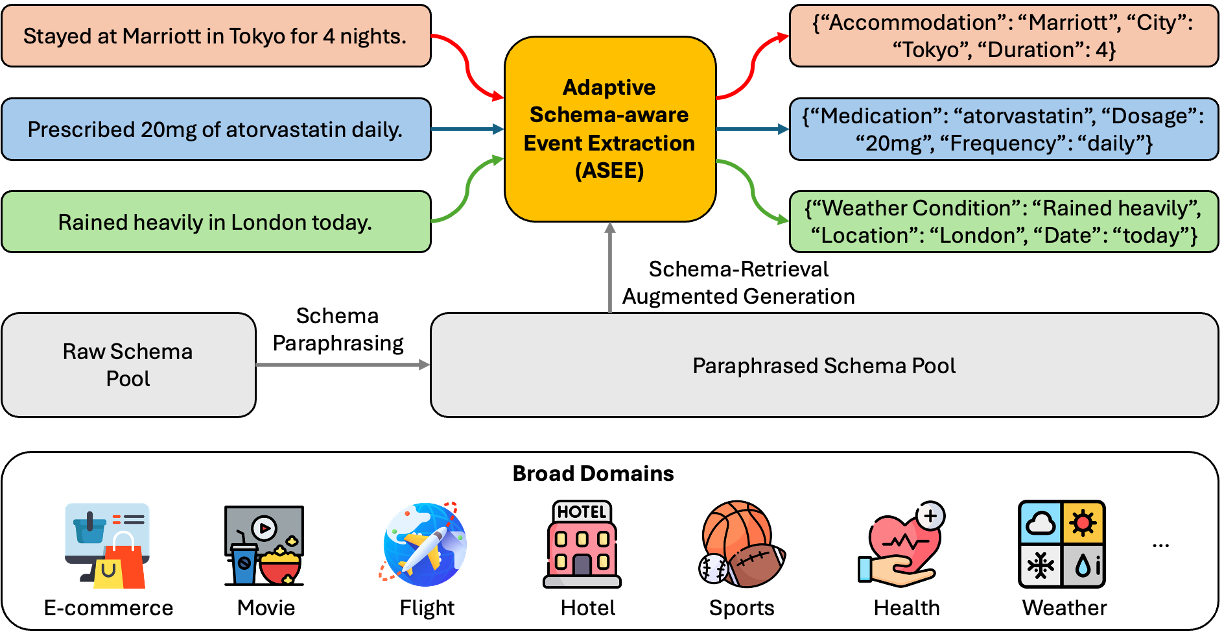}
  \caption{An example of Adaptive Schema-aware Event Extraction (ASEE) in broad domains.}
  \label{fig:example}
  \vspace{-1\baselineskip}
\end{figure}

Event extraction (EE) is an essential task in information extraction (IE)~\cite{xu2023large, lu2022unified} that aims to identify event triggers (i.e., Event Detection (ED)) and their associated arguments (i.e., Event Arguments Extraction (EAE)) from unstructured text, thereby transforming raw text into structured event representations~\cite{xu2023large}.
Event extraction plays an important role in various natural language processing applications, such as knowledge graph construction, question answering, information retrieval, and event prediction, by providing structured representations of real-world occurrences~\cite{lai2022event}.

However, practical EE implementations reveal significant discrepancies from academic assumptions. Real-world deployment necessitates a two-stage paradigm: practitioners must first select appropriate schemas from hundreds of domain-specific candidates before executing extraction with chosen templates. This operational requirement exposes two fundamental limitations in current research.

\textbf{First}, existing pipeline systems suffer from inflexible schema adherence. Current EE pipeline systems operate under a paradoxical assumption: they either rigidly fix event schemas during deployment or naively concatenate all available schemas. The former approach fails in cross-domain scenarios, while the latter causes schema conflicts and error propagation when handling overlapping event types. This rigidity severely limits real-world applicability. Though large language models (LLMs) initially demonstrate promise through their generalization capabilities~\cite{huang2025survey}, two critical shortcomings emerge: (1) Schema hallucination where LLMs invent non-existent event types~\cite{huang2025survey}, and (2) Context window constraints that prevent loading all schema candidates as prompts~\cite{naveed2023comprehensive,wang2024beyond}. While retrieval-augmented generation (RAG) methods~\cite{gao2023retrieval} partially address these issues, current implementations either exceed context capacity with full schema loads~\cite{shiri-2024-decompose} or use oversimplified schema definitions that hinder LLM comprehension.

\textbf{Second}, the absence of joint evaluation benchmarks creates an academic-industrial disconnect. While recent studies explore LLM-based extraction~\cite{shiri-2024-decompose}, existing datasets~\cite{lai2022event} presuppose perfect schema selection—an assumption invalid in real-world scenarios where schema retrieval constitutes a mission-critical prerequisite. This discrepancy leaves the combined capability of schema selection and event extraction unevaluated.

To address the first limitation, we propose  \textbf{A}daptive \textbf{S}chema-aware \textbf{E}vent \textbf{E}xtraction (\textbf{ASEE}), a novel paradigm integrating schema paraphrasing with schema retrieval-augmented generation, by decomposing the event extraction task into schema retrieval and schema-aware extraction.
Figure~\ref{fig:example} shows an example of ASEE in broad domains.
In particular, ASEE extensively builds event extraction schemas, adaptively retrieves specific schemas, automatically assembles event extraction prompts, and accurately generates targeted structures. 

To resolve the second limitation, we construct thewe developed the \textbf{M}ulti-\textbf{D}imensional \textbf{S}chema-aware \textbf{E}vent \textbf{E}xtraction (\textbf{MD-SEE}) benchmark by systematically consolidating 12 datasets, enabling rigorous joint evaluation on schema selection and extraction across various domains, complexities, and language settings.
This benchmark enables rigorous joint assessment of schema selection and extraction accuracy, addressing the current evaluation vacuum.
Comprehensive evaluation show that our proposed ASEE method demonstrates strong adaptability across various scenarios and further enhances event extraction accuracy. 

Our principal contributions are threefold: 
\begin{itemize}
\item \textbf{Adaptive Schema-aware EE Framework}: We propose ASEE, the first framework that jointly enhances the event extraction capability through LLM-based schema paraphrasing and schema-retrieval augmented generation.
\item \textbf{Multi-Dimensional EE Benchmark}: We construct MD-SEE, the first benchmark evaluating both schema matching and extraction accuracy across domains, complexity, and language settings.
\item \textbf{Empirical Insights}: Through extensive experiments and analysis, we provide insightful results for event extraction with LLMs across broad domains, providing actionable guidelines for industrial deployment.
\end{itemize}

\section{Related Work}

\paragraph{Information Extraction Task}

Information extraction (IE) aims to automatically extract structured knowledge from unstructured texts, typically involving three core tasks: (1) Named Entity Recognition (NER) for identifying entity boundaries and types~\cite{ye2024llm}, (2) Relation Extraction (RE) for detecting semantic relationships between entities~\cite{wang2304instructuie}, and (3) Event Extraction (EE) for recognizing event triggers and their associated arguments~\cite{wang2022deepstruct}. These tasks are conventionally categorized as closed IE (with predefined schemas) or open IE (schema-agnostic). Closed IE relies on predefined schemas specifying target entities, relations, or event structures, enabling precise extraction in constrained domains~\cite{lu2022unified}. Open IE systems, conversely, extract open-domain triples without schema constraints, sacrificing precision for broader coverage~\cite{wang2304instructuie}.

Existing event extraction methods predominantly follow the closed IE paradigm, requiring predefined schemas that specify event types, roles, and constraints. Though effective in controlled settings, such rigid schemas face two key challenges: (1) \textit{Oversimplification} - Most schemas abstract away domain-specific nuances (e.g., using generic role labels like "participant"), making them insufficient for guiding LLMs in real-world extraction; (2) \textit{Semantic Gap} - Predefined schemas often mismatch actual context semantics, especially when processing cross-domain or emerging event types.

This paper proposes Adaptive Schema-aware Event Extraction (ASEE) to address real-world EE scenarios requiring schema selection before extraction. Unlike traditional closed EE with rigid predefined schemas, our framework dynamically adapts schemas through : (1) Schema Paraphrasing that rewrites schema elements using contextual knowledge, and (2) Schema-Retrieval Augmented Generation guided by retrieving the relevant schemas and extracting events with the matched schemas.

\paragraph{Event Extraction with LLMs}

Recently, a significant amount of work has focused on utilizing large language models (LLMs) to implement and enhance the performance of information extraction tasks.
These works can be primarily divided into four aspects:
(1) data augmentation: leveraging LLMs to generate synthetic data or augment existing datasets for information extraction tasks, avoiding the introduction of unrealistic, misleading, and offset patterns~\cite{wang2023improving}.
(2) prompt engineering: crafting effective prompts for LLMs to guide them in extracting relevant information from text~\cite{ashok2305promptner}.
(3) code LLMs: utilizing code-based LLMs to automate and enhance the information extraction process~\cite{guo2023retrieval}.
(4) LLM fine-tuning: fine-tuning LLMs on domain-specific data to optimize their performance on targeted information extraction tasks~\cite{zhou2023universalner}.

However, when using LLMs for event extraction (EE), they tend to experience significant hallucinations due to insufficient task-specific optimization, and their limited context window makes it impractical to rely solely on LLMs to handle event extraction tasks across diverse scenarios.
Retrieval-augmented generation (RAG) methods have emerged as a promising solution, enhancing extraction accuracy by incorporating external knowledge~\cite{gao2023retrieval}.
For instance, \citet{shiri-2024-decompose} decompose event extraction into two subtasks: Event Detection (ED), which retrieves relevant event examples, and Event Argument Extraction (EAE), which extracts events based on the retrieved examples.
In this paper, we aim to mitigate the hallucination issue of LLMs in EE by adopting a schema-retrieval augmented generation method, while also addressing the issue of LLM context length limitations.

\section{Methodology}
\label{sec:methodology}

\subsection{Preliminaries}

Schema-aware Event Extraction (SEE)~\cite{shiri-2024-decompose,gui-etal-2024-iepile} leverages predefined schemas to guide the extraction of structured information from unstructured text. A schema serves as a formal representation of the types of information to be extracted, encompassing entities, relationships, events, and their attributes. By defining the structure and constraints of the desired data, schemas enable the extraction system to identify and organize relevant information systematically.

Given a query text $q$, which can range from a single sentence to an entire document, and a schema $s$ that specifies a particular type of information to be extracted along with its associated arguments, the goal of SEE is to extract relevant information from $q$ according to $s$. Each schema $s$ consists of a set of arguments $\mathcal{A} = \{a_{1}, a_{2}, \dots, a_{K}\}$, where $K$ is the number of arguments in schema $s$.

Formally, the Schema-aware Event Extraction (SEE) task can be represented as $\mathcal{V} = \theta(s, q)$, where $\mathcal{V} = \{v_{1}, v_{2}, \dots, v_{K}\}$ denotes the set of extracted values corresponding to the arguments in schema $s$, and $\theta$ represents the extraction model that takes schema $s$ and query $q$ as inputs to produce the extracted information.

SEE assumes the availability of predefined schemas, which may not always be feasible in real-world scenarios. Acquiring accurate and comprehensive schemas from a wide range of domains and languages can be resource-intensive and challenging. Our proposed Adaptive Schema-aware Event Extraction (ASEE) framework addresses this issue by incorporating schema paraphrasing and retrieval, enabling robust and versatile event extraction across diverse scenarios.

\subsection{Adaptive Schema-aware EE (ASEE)}

\begin{figure*}[htbp]
  \centering
  \includegraphics[width=1.0\textwidth]{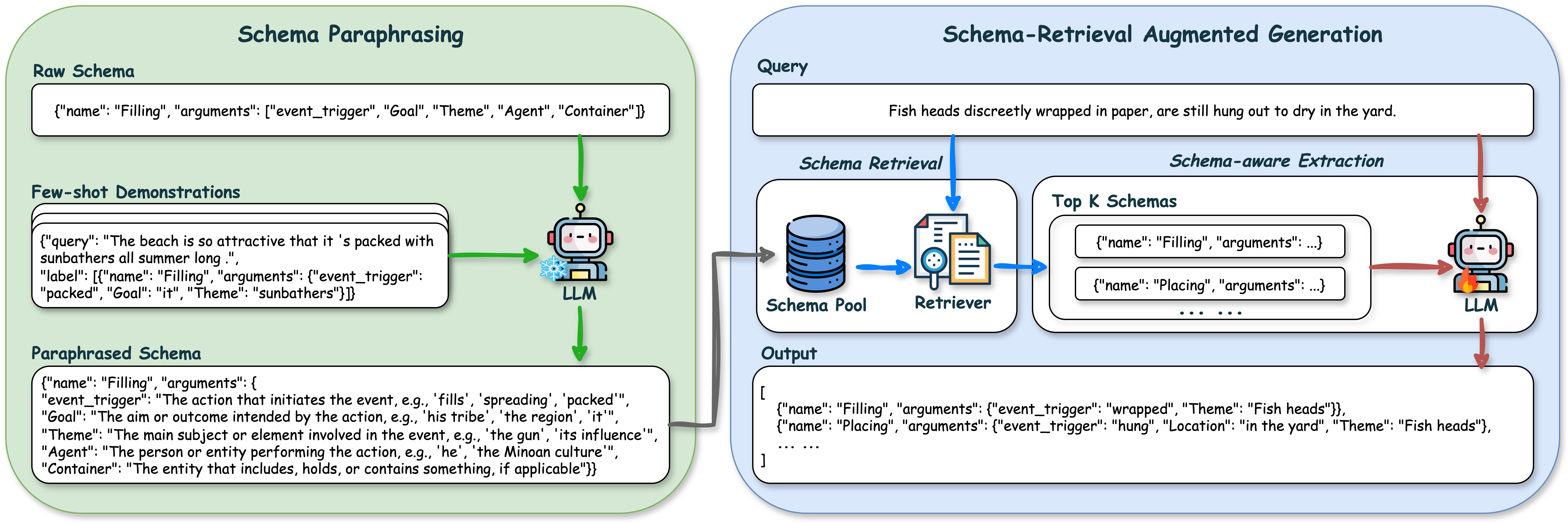}
  \caption{The architecture of ASEE comprises two primary components: Schema Paraphrasing (SP) and Schema-Retrieval Augmented Generation (SRAG, including Schema Retrieval (SR) and Schema-aware Extraction (SE)).}
  \label{fig:task}
  \vspace{-1\baselineskip}
\end{figure*}

Building upon the foundational concepts of SEE~\cite{shiri-2024-decompose,gui-etal-2024-iepile}, we introduce our proposed framework, Adaptive Schema-aware Event Extraction (ASEE). ASEE addresses the inherent challenges of SEE by incorporating adaptive mechanisms for schema retrieval and extraction. The framework is visually represented in Figure~\ref{fig:task} and comprises two primary components: Schema Paraphrasing (SP) and Schema-Retrieval Augmented Generation (SRAG, including Schema Retrieval (SR) and Schema-aware Extraction (SE)).

The ASEE framework can be formally represented by the following sequence of operations, with each step explained alongside its corresponding formula:

\noindent$\bullet$ \textbf{Step 1 - Schema Paraphrasing (SP):}
\begin{align}
    \mathcal{S} &= \bigcup_{s \in \mathcal{S}_0}\{\phi_{\text{LLM}}\left(s, D_{s}\right)\left|D_{s} \subseteq \mathcal{D}_{\text {train }}\}\right. \label{eq:schema_paraphrasing}
\end{align}
Equation~\ref{eq:schema_paraphrasing} represents the schema paraphrasing process. Here, $\phi_{\text{LLM}}$ denotes the schema paraphrasing function, which takes each initial schema $s$ from the set $\mathcal{S}_0$ and generates a paraphrased schema pool $\mathcal{S}$ using a subset of the training data $D_s$ as few-shot examples.

\noindent$\bullet$ \textbf{Step 2 - Schema Retrieval (SR):}
\begin{align}
    \mathcal{R}_q &= \psi_{\text{retriever}}(q, \mathcal{S}) \label{eq:schema_retrieval}
\end{align}
Equation~\ref{eq:schema_retrieval} describes the schema retrieval step. The function $\psi_{\text{retriever}}$ is the retrieval model that takes the query text $q$ and the paraphrased schema pool $\mathcal{S}$ to retrieve the top-$k$ relevant schemas, resulting in the set $\mathcal{R}_q$.

\noindent$\bullet$ \textbf{Step 3 - Schema-aware Extraction (SE):}
\begin{align}
    \mathcal{V} &= \theta_{\text{LLM}}(q, \mathcal{R}_q) \label{eq:information_extraction}
\end{align}
Equation~\ref{eq:information_extraction} outlines the information extraction process. The function $\theta_{\text{LLM}}$ represents the LLM fine-tuned for extraction, which takes the query $q$ and the retrieved schemas $\mathcal{R}_q$ to produce the final set of extracted values $\mathcal{V}$.

\subsubsection{Schema Paraphrasing (SP)}

Schema Paraphrasing (SP) serves as the preparation stage and backbone for the extraction process, establishing a robust and comprehensive schema pool.
For a given event extraction task, we first collect all relevant schemas $\mathcal{S}_0$. For each schema $s \in \mathcal{S}_0$, we utilize data samples from the training set that adhere to schema $s$ as few-shot demonstrations to guide a frozen LLM in generating paraphrased versions of the original schema. These paraphrased schemas introduce detailed argument descriptions, enhancing the semantic clarity of the schemas and facilitating more effective retrieval. The resulting paraphrased schemas are aggregated to form the schema pool $\mathcal{S}$, which serves as a repository of diverse schemas tailored for various extraction tasks, as defined in Equation~\ref{eq:schema_paraphrasing}.

\subsubsection{Schema-Retrieval Augmented Generation (SRAG)}

The extraction component encompasses the operational aspects of the ASEE framework, divided into Schema Retrieval (SR) and Schema-aware Extraction (SE) processes.

\paragraph{Schema Retrieval (SR)}

Upon receiving a new query text $q$, the first task is to identify the most relevant schemas from the schema pool $\mathcal{S}$. This is achieved through a schema retrieval mechanism that employs a specialized retrieval model $\psi$. The retriever processes the query $q$ to search the schema pool and retrieves the top-$k$ schemas $\mathcal{R}_q = \{s_1, s_2, \dots, s_k\}$ that are most pertinent to the information contained within $q$, as defined in Equation~\ref{eq:schema_retrieval}. This retrieval step ensures that the subsequent extraction is guided by schemas that are contextually aligned with the query, enhancing extraction accuracy and relevance.

\paragraph{Schema-aware Extraction (SE)}  
With the relevant schemas $\mathcal{R}_q$ identified, the next step involves the information extraction process. While large language models (LLMs) can directly perform extraction based on schemas, their zero-shot performance may be suboptimal due to challenges in strictly adhering to schema constraints or interpreting complex argument definitions. To address this, we employ \textbf{Supervised Fine-Tuning (SFT)} on the training set to adapt the base LLM for schema-guided extraction. 

Given a dataset $\mathcal{D}_{\text{train}} = \{(q_i, s_i, \mathcal{V}_i)\}_{i=1}^N$ containing query texts, schemas, and ground-truth structured outputs, we fine-tune the LLM to minimize the discrepancy between its predictions and the target values. Formally, for each sample $(q, s, \mathcal{V})$, the model $\theta$ is optimized to maximize the likelihood of generating the correct argument values conditioned on the input query $q$ and schema $s$. The SFT objective is defined as:  
\begin{align}
  \mathcal{L}_{\text{SFT}} &= -\mathbb{E}_{(q, s, \mathcal{V}) \sim \mathcal{D}_{\text{train}}} \sum_{k=1}^K \log P_\theta\left(v_k \mid q, s, v_{<k}\right), \label{eq:sft}
\end{align}  
where $v_k$ denotes the $k$-th argument value in $\mathcal{V}$, and $v_{<k}$ represents previously generated values. This autoregressive training enables the model to learn schema-specific dependencies and formatting constraints. After fine-tuning, the LLM $\theta_{\text{LLM}}$ in Equation~\ref{eq:information_extraction} becomes specialized in generating structured outputs that strictly comply with the retrieved schemas $\mathcal{R}_q$. The final output $\mathcal{V}$ comprises structured information extracted from $q$, organized according to the specifications of $\mathcal{R}_q$.  

\section{Benchmark Constrcution}~\label{sect:dataset}
\vspace{-1\baselineskip}

We construct the \textbf{M}ulti-\textbf{D}imensional \textbf{S}chema-aware \textbf{E}vent \textbf{E}xtraction (\textbf{MD-SEE}) benchmark to facilitate rigorous evaluation, which systematically consolidates 12 datasets across diverse domains, complexity levels, and language settings.
Due to space constraints, additional details are provided in the Appendix~\ref{app:data_collection}.

\subsection{Data Collection}
\label{sec:datasets}
Datasets were collected from various sources, focusing on those that are fully open-source to avoid licensing restrictions. The collected datasets include DocEE\cite{tong-etal-2022-docee}, 
{MAVEN-Arg}\cite{wang-etal-2024-maven}, 
{GENEVA}\cite{parekh-etal-2023-geneva},
{CrudeOilNews}\cite{lee-etal-2022-crudeoilnews},
and the event extraction datasets from IEPILE\cite{gui-etal-2024-iepile}, which contains out-of-domain testing tests to evaluate the generalize capability. Table~\ref{tab:dataset_statistics_1} in Appendix~\ref{app:data_collection} provides a summary of the collected datasets.
We perform initial cleaning to ensure data quality, detailed procedures are provided in Appendix~\ref{app:schema_processing} for schema processing and Appendix~\ref{app:dataset_processing} for dataset processing.

\begin{figure}[t]
\centering
  \includegraphics[width=\columnwidth]{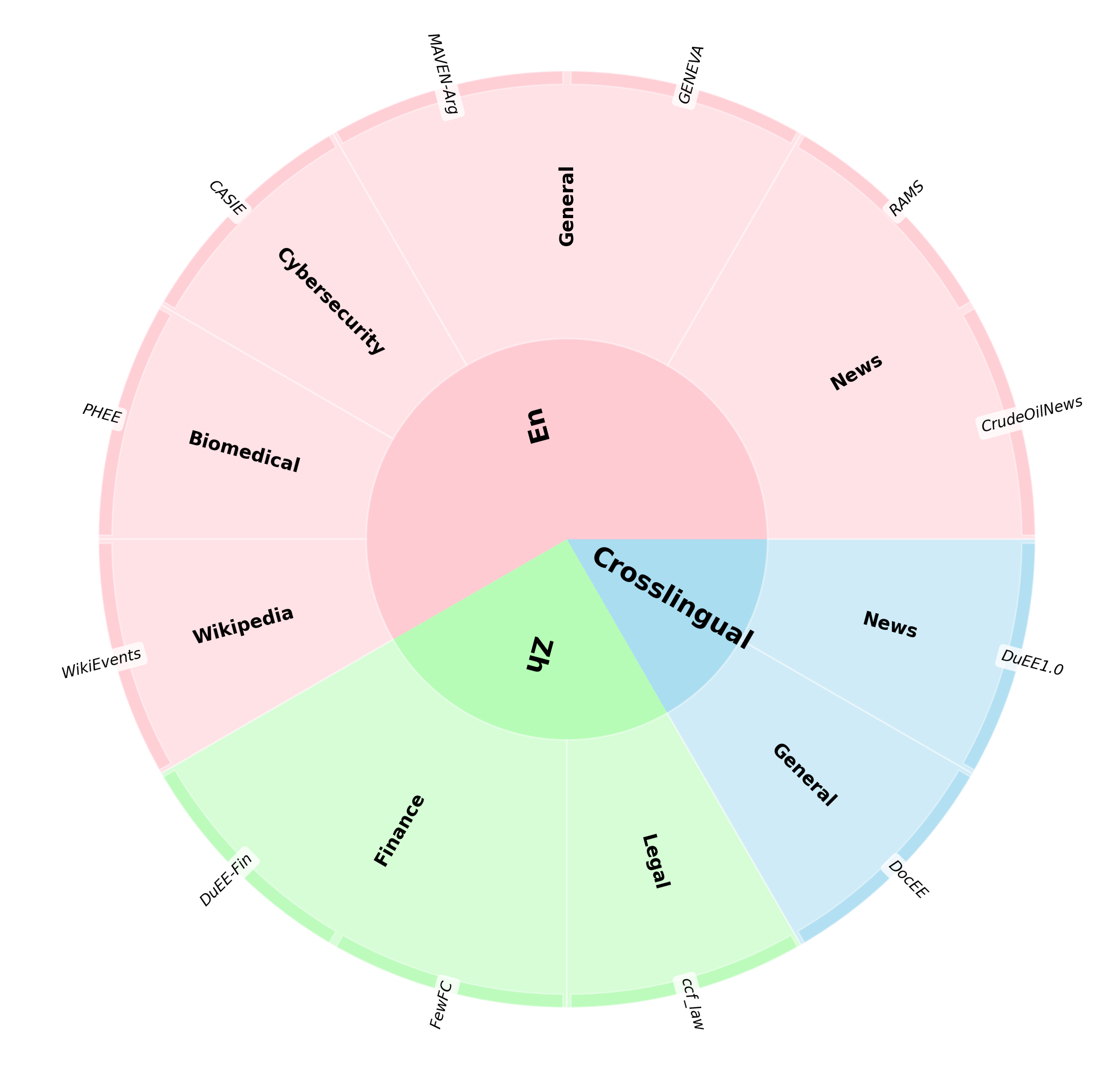}
  \caption{Data Statistics of MD-SEE.}
  \label{fig:MD-SEE}
  \vspace{-1\baselineskip}
\end{figure}

\subsection{MD-SEE Dataset}

To demonstrate the capabilities of {ASEE}, we developed the \textbf{M}ulti-\textbf{D}imensional \textbf{S}chema-aware \textbf{E}vent \textbf{E}xtraction (\textbf{MD-SEE}) dataset. Constructed by aggregating the collected datasets, MD-SEE ensures comprehensive coverage across a wide range of non-overlapping schemas. The multidimensional nature of MD-SEE is characterized by:

\noindent$\bullet$ \textbf{Various Query Lengths}: Supports extraction from sentence-level to document-level queries, enabling evaluation across different textual granularities.

\noindent$\bullet$ \textbf{Multiple Domains}: Incorporates datasets from diverse domains such as news, cybersecurity, biomedical, finance, and legal sectors, ensuring generalization across varied contexts.

\noindent$\bullet$ \textbf{Diverse Event Complexity}: Accommodates both single-event and multi-event extraction scenarios, testing the system's adaptability in handling complex event structures.

\noindent$\bullet$ \textbf{Multiple Language Settings}: Encompasses English-only, Chinese-only, and cross-lingual extraction subsets, highlighting proficiency in multilingual and cross-lingual information extraction.

\subsubsection{Schema Consolidation}

To enhance schema diversity and reduce redundancy, we performed schema consolidation. We merged all schemas from individual datasets and conducted a manual merging step to unify nearly identical schemas and cross-lingual duplicates (English and Chinese). Detailed procedures are the same as in Appendix~\ref{app:schema_processing}.

Subsequently, we used the multilingual sentence embedding model BGE-M3~\cite{bgem3}  to encode schemas. We constructed a graph where each node represents a schema, and edges connect schemas with cosine similarity above 0.85. Applying a Greedy Maximum Independent Set algorithm (Algorithm~\ref{alg:greedy_max_indep_set} in Appendix~\ref{app:GMIS}), we identified the largest possible subset of diverse schemas by removing highly similar ones. We then filtered the dataset to include only the samples associated with these schemas, ensuring relevance and alignment within MD-SEE.

\subsubsection{Cross-Lingual Subset}

To incorporate cross-lingual extraction capabilities, we processed subsets from {DocEE} and {DuEE1.0}:

\noindent$\bullet$ \textbf{DocEE}: After consolidation, only English queries remained. We translated all schemas from English to Chinese and adjusted the labels, resulting in English queries paired with Chinese schemas.

\noindent$\bullet$ \textbf{DuEE1.0}: Originally in Chinese, we translated schemas from Chinese to English and modified the labels, producing Chinese queries associated with English schemas.

These cross-lingual subsets require the model to interpret schemas in one language while extracting information from queries in another, preserving the extracted values in the query's original language. This setup evaluates the model's ability to generalize across languages and handle multilingual scenarios common in real-world applications.

\subsubsection{Dataset Statistics}

Shown as Figure~\ref{fig:MD-SEE}, the consolidated MD-SEE dataset comprises {300} schemas covering multiple dimensions, {12,817} training samples, {1,775} development samples, and {7,686} test samples. Detailed statistics are provided in Table~\ref{tab:dataset_statistics_2} (Appendix~\ref{app:dataset_statistics}). This aggregation ensures that MD-SEE is robust and versatile, catering to various event extraction tasks across different contexts.

\section{Experiments}
\subsection{Evaluation Settings}

Our evaluation considers three main aspects: 1) schema retrieval evaluation, 2) schema-aware extraction evaluation, and 3) end-to-end evaluation, providing a comprehensive assessment across various dimensions.

\subsubsection{Schema Retrieval Evaluation}
Given a query ranging from a sentence to a full document, the schema retrieval task expects to retrieve the most relevant schemas from a schema pool.

\paragraph{Metric}
We use \textbf{Recall@K} to measure the proportion of ground-truth schemas included in the top-$K$ retrieved schemas for each query. A high Recall@K indicates effective identification of potential schemas.

\subsubsection{Schema-aware Extraction Evaluation}
Provided with both the query $q$ and the corresponding ground-truth schemas $S_q$, the objective of the extraction evaluation is to extract information accurately according to the paraphrased schemas.

\paragraph{Metric}
We evaluate extraction performance by calculating the \textbf{F1} score for each argument within each schema and then averaging these scores over all arguments and schemas.

\subsubsection{End-to-End Evaluation}
The end-to-end evaluation assesses the system's performance in autonomously retrieving relevant schemas and extracting information based on the matched schemas without prior knowledge of the ground-truth schemas. For each query, the system:

\begin{itemize}
    \item \textbf{Schema Retrieval}: Retrieve a set of schemas $R_q$ based on the query $q$.
    \item \textbf{Schema-aware Extraction}: Extract information from the query $q$ according to retrieved schemas $R_q$.
\end{itemize}

\paragraph{Metric}
We use a modified End-to-End F1 Score (\textbf{E2E-F1}), defined as:
\begin{equation}
F1 = \frac{1}{N|S_q|} \sum_{q=1}^{N} \sum_{s \in S_q} 
    F1(s, q) \cdot \mathbb{I}(s \in R_q)
\end{equation}
Here, $\mathbb{I}(s \in R_q)$ is an indicator function that equals 1 if schema $s$ is retrieved for query $q$, and 0 otherwise. The E2E-F1 considers the following cases:
\noindent$\bullet$ \textbf{Schema Retrieved and Relevant}: If a ground-truth schema $s \in S_q$ is retrieved ($s \in R_q$), we evaluate the extraction for that schema using $F1(s, q)$.

\noindent$\bullet$ \textbf{Schema Not Retrieved}: If a ground-truth schema $s \in S_q$ is not retrieved ($s \notin R_q$), we assign an F1 score of zero for that schema.

\noindent$\bullet$ \textbf{Schema Retrieved but Not Relevant}: Retrieved schemas not in the ground truth ($s \in R_q$, $s \notin S_q$) are ignored in the evaluation.

This evaluation framework mirrors real-world scenarios where both retrieval and extraction accuracy are critical, emphasizing the importance of effectively identifying and extracting relevant information.

\subsection{Experimental Settings}

\begin{table*}[ht]
    \centering
    \footnotesize
    \addtolength{\tabcolsep}{-4pt}

    \begin{tabular}{l|cc|cc|cc|cc|cc|cc|cc}
        \toprule
        & \multicolumn{2}{c|}{CrudeOilNews} & \multicolumn{2}{c|}{DocEE-en} & \multicolumn{2}{c|}{DocEE-zh} & \multicolumn{2}{c|}{GENEVA} & \multicolumn{2}{c|}{IEPILE-en} & \multicolumn{2}{c|}{IEPILE-zh} & \multicolumn{2}{c}{MAVEN-Arg} \\
        & Raw & Paraph. & Raw & Paraph. & Raw & Paraph. & Raw & Paraph. & Raw & Paraph. & Raw & Paraph. & Raw & Paraph. \\
        \midrule
        BM25 & \textbf{0.35} & 0.31 & 0.21 & \textbf{0.67} & 0.55 & \textbf{0.77} & 0.11 & \textbf{0.41} & 0.61 & \textbf{0.78} & 0.75 & \textbf{0.81} & 0.09 & \textbf{0.22} \\
        BGE-M3 & \textbf{0.35} & 0.32 & 0.85 & \textbf{0.92} & 0.73 & \textbf{0.94} & 0.43 & \textbf{0.50} & 0.77 & \textbf{0.88} & 0.84 & \textbf{0.91} & 0.13 & \textbf{0.33} \\
        E5-LV2 & 0.35 & \textbf{0.36} & 0.84 & \textbf{0.87} & 0.41 & \textbf{0.51} & 0.43 & \textbf{0.51} & 0.83 & \textbf{0.90} & \textbf{0.46} & 0.30 & 0.20 & \textbf{0.31} \\
        GTE-L & 0.33 & \textbf{0.37} & 0.95 & 0.95 & 0.47 & \textbf{0.50} & 0.40 & \textbf{0.48} & 0.84 & \textbf{0.89} & 0.34 & \textbf{0.50} & 0.19 & \textbf{0.32} \\
        LLM-E & 0.30 & \textbf{0.31} & 0.88 & \textbf{0.90} & \textbf{0.34} & 0.33 & 0.40 & \textbf{0.51} & 0.82 & \textbf{0.88} & 0.27 &\textbf{ 0.53} & 0.17 & \textbf{0.30} \\
        BGE-RB & 0.31 & \textbf{0.33} & 0.43 & \textbf{0.49} & \textbf{0.48} & 0.44 & 0.33 & \textbf{0.47} & 0.66 & \textbf{0.83} &\textbf{ 0.89} & 0.87 & 0.10 & \textbf{0.16} \\
        BGE-RL & 0.28 & \textbf{0.31} & 0.57 & \textbf{0.77} & 0.66 & \textbf{0.69} & 0.41 & 0.41 & 0.78 & \textbf{0.81} & 0.89 & \textbf{0.90} & 0.14 & \textbf{0.20} \\
        \bottomrule
    \end{tabular}
    
    \caption{Schema retrieval evaluation results on individual datasets, using Recall@10 as the metric. Better results are highlighted in \textbf{bold}.}
    \label{tab:retrieval_results_2}
\end{table*}

\begin{table}[ht]
\footnotesize
\centering
\addtolength{\tabcolsep}{-3pt}
    \begin{tabular}{l|cc|cc|cc}
        \toprule
        & \multicolumn{2}{c|}{Recall$@$10} & \multicolumn{2}{c|}{Recall$@$20} & \multicolumn{2}{c}{Recall$@$50} \\
        & Raw & Paraph. & Raw & Paraph. & Raw & Paraph. \\
        \midrule
        BM25 & 0.33 & \textbf{0.58} & 0.39 & \textbf{0.67} & 0.49 & \textbf{0.76} \\
        BGE-M3 & 0.61 & \textbf{0.78} & 0.69 & \textbf{0.86} & 0.78 & \textbf{0.94} \\
        E5-LV2 & 0.35 & \textbf{0.61} & 0.43 & \textbf{0.72} & 0.62 & \textbf{0.86} \\
        GTE-L & 0.28 & \textbf{0.57} & 0.36 & \textbf{0.68} & 0.51 & \textbf{0.82} \\
        LLM-E & 0.34 & \textbf{0.71} & 0.49 & \textbf{0.82} & 0.68 & \textbf{0.91} \\
        BGE-RB & 0.51 & \textbf{0.59} & 0.60 & \textbf{0.66} & 0.70 & \textbf{0.76} \\
        BGE-RL & 0.56 & \textbf{0.61} & 0.64 & \textbf{0.69} & 0.75 & \textbf{0.79} \\
        \bottomrule
    \end{tabular}
    \caption{Schema retrieval evaluation results on MD-SEE. Better results are highlighted in \textbf{bold}.}
    \label{tab:retrieval_results_1}
\end{table}

\paragraph{Retrieval Models.}
We employed the following seven retrieval models, which are commonly used in RAG scenarios and support multiple languages, including:
{BM25}~\cite{DBLP:journals/ftir/RobertsonZ09},
{BGE-M3}, {BGE-Reranker-Base} (BGE-RB), and {BGE-Reranker-Large} (BGE-RL)~\cite{bgem3},
{E5-large-v2} (E5-LV2)~\cite{Wang2022TextEB},
{GTE-Large} (GTE-L) \cite{Li2023TowardsGT}
{LLM-Embedder} (LLM-E)~\cite{zhang-etal-2024-multi-task}.

\paragraph{LLMs.}
Within the limits of our computational resources, we employed the following state-of-the-art large language models and information extraction models to carry out our information extraction tasks, including:
{Phi-3.5-mini}~\cite{Abdin2024Phi3TR}, {Llama-3.2-3B} and {Llama-3.1-8B}~\cite{Dubey2024TheL3}, {Mistral-7B-v0.3}~\cite{jiang2023mistral7b}, {Qwen2.5-7B} and {Qwen2.5-14B}~\cite{Yang2024Qwen2TR}, {YAYI-UIE}~\cite{Xiao2023YAYIUIEAC}, {GPT-4-turbo}~\cite{Wada2024OptimizingGT}

\paragraph{Datasets} We conduct experiments on several collected datasets, including MAVEN-Arg, GENEVA, CrudeOilNews, DocEE (-zh \& -en), IEPILE (-zh \& -en), and our newly developed MD-SEE dataset. These datasets span multiple scenarios, providing comprehensive insights into our ASEE framework.

\subsection{Schema Retrieval Evaluation}
To verify the effectiveness of our proposed ASEE method in improving retrieval performance by paraphrasing the raw schema, we conducted the following experiments.
First, we performed the retrieval experiments to the collected individual datasets, computing Recall$@$10 for both raw schema (denoted as ``Raw'') and schema paraphrasing (denoted as ``Paraph.'').
In most cases presented in Table~\ref{tab:retrieval_results_2}, the ``Paraph.'' consistently showed marked improvements over ``Raw'' across different retrieval models. 

As an addition, we calculated the Recall$@$10, Recall$@$20, and Recall$@$50 for both the raw schema and the schema paraphrasing on the MD-SEE dataset. The results, as shown in Table~\ref{tab:retrieval_results_1}, indicate that ``Paraph.'' outperforms ``Raw'' across all metrics—Recall$@$10, Recall$@$20, and Recall$@$50—for all retrieval models. 
These experiments collectively demonstrate that paraphrasing the raw schema using our proposed few-shot demonstration paraphrasing can effectively enhance schema retrieval performance.

\begin{table*}[ht]
    \centering
    \footnotesize
    \addtolength{\tabcolsep}{-2pt}

    \begin{tabular}{lccccccc}
        \toprule
        & CrudeOilNews & DocEE-en & DocEE-zh & GENEVA & IEPILE-en & IEPILE-zh & MAVEN-Arg \\
        \midrule
        Phi-3.5-mini& 0.22 & 0.43 & 0.46 & 0.31 & 0.33 & 0.60 & 0.29 \\
        Llama-3.2-3B & 0.35 & 0.47 & 0.60 & 0.47 & 0.44 & 0.71 & 0.37 \\
        Llama-3.1-8B & \underline{0.43} & 0.48 & 0.61 & 0.58 & \underline{0.55} & 0.71 & 0.51 \\
        Mistral-7B-v0.3 & 0.37 & 0.45 & 0.57 & 0.57 & 0.49 & 0.71 & 0.47 \\
        Qwen2.5-7B & 0.42 & 0.45 & 0.41 & \underline{0.60} & \underline{0.55} & 0.73 & \underline{0.48} \\
        Qwen2.5-14B & 0.30 & \underline{0.49} & \textbf{0.62} & 0.49 & 0.53 & \textbf{0.78} & 0.40 \\
        YAYI-UIE &0.35 & 0.37	&0.30	&0.43	&0.41	&0.34	&0.33
\\
        GPT-4-turbo&\textbf{0.50} &\textbf{0.56}	&\textbf{0.62}	&\textbf{0.67}	&\textbf{0.62}	&\underline{0.77}	&\textbf{0.56}
\\
        \bottomrule
    \end{tabular}

    \caption{Zero-shot schema-aware extraction results, using F1 score as the metric.The best results are marked in \textbf{bold},while the second \underline{underlined}.}
    \label{tab:extraction_results_1}
\end{table*}

\begin{table*}[ht]
\centering
\footnotesize
\addtolength{\tabcolsep}{2pt}
\begin{tabular}{lccccccc}
    \toprule
    & BM25 & BGE-M3 & E5-LV2 & GTE-LG & LLM-E & BGE-RB & BGE-RL \\
    \midrule
    Llama-3.2-3B & 0.46 & 0.62 & 0.48 & 0.46 & 0.57 & 0.47 & 0.48 \\
    w/ SFT & \textbf{0.47} & \textbf{0.63} & \textbf{0.49} & \textbf{0.47} & \textbf{0.58} & \textbf{0.48} & \textbf{0.49} \\
    \midrule
    Llama-3.1-8B         & 0.48 & 0.65 & 0.51 & 0.47 & 0.59 & 0.49 & 0.51 \\
    w/ SFT               & \textbf{0.50} & \textbf{0.68} & \textbf{0.53} & \textbf{0.50} & \textbf{0.62} & \textbf{0.51} & \textbf{0.53} \\
    \bottomrule
\end{tabular}
\caption{End-to-end results of MD-SEE across different retrieval and extraction model configurations, using E2E-F1 score as the metric. Better results are highlighted in \textbf{bold}.}
\label{tab:End to end results of MD-SEE}
\end{table*}

\subsection{Schema-aware Extraction Evaluation}

To evaluate the performance of state-of-the-art LLMs for event extraction, we conducted experiments on zero-shot schema-aware event extraction tasks on multiple datasets using the ground-truth paraphrased schemas. As shown in Table~\ref{tab:extraction_results_1}, we used six state-of-the-art open-source LLMs (i.e., Phi-3.5-mini, Llama-3.2-3B, Llama-3.1-8B, Mistral-7B-v0.3, Qwen2.5-7B, Qwen2.5-14B), an information extraction model based on an LLM (i.e., YAYI-UIE), and the popular closed-source model GPT-4 (i.e., GPT-4-turbo). 
Notably, we \emph{bypassed the schema matching step and directly provided each extraction task with an optimal event extraction schema}. 

The results clearly indicate that GPT-4 has the strongest zero-shot event extraction capabilities, outperforming all other models on most datasets. However, in the Chinese data sets DocEE-zh and IEPILE-zh, it was closely matched by Qwen2.5-14B, which specializes in Chinese. We also observed that YAYI-UIE, a universal information extraction method, did not achieve satisfactory event extraction results compared to the open-source LLM-based methods. Despite having 14 billion parameters, its performance lagged behind due to the limited capabilities of its backbone LLM, performing even worse than Llama-3.2-3B.

\subsection{End-to-End Evaluation}
To comprehensively assess the performance of our ASEE framework, we conducted end-to-end evaluations. 
Table~\ref{tab:End to end results of MD-SEE} shows the end-to-end results of MD-SEE across different retrieval and extraction model configurations, using E2E-F1
score as the metric.
This evaluation encompasses both schema retrieval and information extraction components, reflecting real-world application scenarios where the system must autonomously retrieve relevant schemas and accurately extract information without prior knowledge of ground-truth schemas.

Table~\ref{tab:End to end results of MD-SEE} shows that ASEE, using BGE-M3 as the schema retriever, achieves the best overall performance in E2E-F1 compared to other retrieval models (e.g., BM25) across both Llama-3.2-3B and Llama-3.1-8B, with and without SFT, for schema-aware event extraction.
The improved retrieval enabled by schema paraphrasing allows ASEE to identify more relevant schemas, thereby providing a stronger foundation for accurate event extraction.

In addition, ASEE with Llama-3.1-8B achieves better E2E-F1 performance than ASEE with Llama-3.2-3B, which is a smaller and less powerful LLM.
The end-to-end event extraction performance of ASEE can be further enhanced when the extraction LLM (e.g., Llama-3.2-3B or Llama-3.1-8B) is trained with supervised fine-tuning (SFT).
These results demonstrate that ASEE benefits from using a larger and/or fine-tuned LLM for schema-aware extraction.

Overall, our proposed ASEE method can be enhanced either by a strong retrieval model for paraphrased schema matching or by a larger and/or fine-tuned LLM for schema-aware extraction.

\section{Conclusion}
In this paper, we introduced Adaptive Schema-aware Event Extraction (ASEE), a novel framework designed to enhance event extraction (EE) across diverse domains using large language models (LLMs). 
By decomposing the extraction process into schema paraphrasing and schema retrieval-augmented extraction, ASEE effectively mitigates challenges such as hallucinations and context length limitations inherent in LLM-based approaches. 
We develop the MD-SEE dataset, which consists of high-quality schemas across various dimensions, providing a comprehensive resource for evaluating event extraction systems. 
We conducted extensive experiments on multiple datasets including our newly developed MD-SEE dataset, demonstrating ASEE's adaptability and superior extraction performance. 
The framework's ability to build and leverage a comprehensive schema pool enables more precise and scalable event extraction, making it a robust solution for a wide range of real-world applications. 
Future work will explore integrating relation extraction and named entity recognition, fine-tuning larger LLMs, and extending ASEE to more complex multilingual and cross-lingual scenarios to further enhance its capabilities.

\section{Limitations}
Our Adaptive Schema-aware Event Extraction (ASEE) framework has several limitations. First, we did not incorporate relation extraction (RE) and named entity recognition (NER) due to our focus on schema-based extraction tasks, for which event extraction is more suited. Second, we were unable to fine-tune larger language models, such as those with 32B or 70B parameters, owing to computational resource constraints. Additionally, our current implementation does not address more complex multilingual and cross-lingual scenarios, which presents further challenges for scalable and versatile information extraction. We hope to address the above limitations in the follow-up work.

\balance

\bibliography{paper}

\clearpage

\nobalance

\appendix

\section{Dataset Collection}
~\label{app:data_collection}

In this section, we present the details of dataset collection, including schema processing (Appendix~\ref{app:schema_processing}), dataset processing (Appendix~\ref{app:dataset_processing}), the schema consolidation algorithm (Appendix~\ref{app:GMIS}), and dataset statistics (Appendix~\ref{app:dataset_statistics}).

\subsection{Schema Processing}~\label{app:schema_processing}

Each dataset was examined to extract unique schemas, which include the schema name, description, arguments, and relevant metadata. We employed a heuristic merging process to address potential duplications. This included:

\begin{enumerate}
    \item \textbf{Character Similarity:} Schemas were merged if their names and arguments shared over 80\% character similarity, thus reducing redundancy due to variations in tense or plural forms.
    \item \textbf{Numerical and Variant Arguments:} For arguments with numerical variants (e.g., place, place1, place2), we consolidated them into a single argument with values stored in a list. This approach enhances the schema's clarity and usability.
\end{enumerate}

These efforts ensure that our schema pool reflects a rich and diverse set of scenarios, ultimately enhancing the robustness of the data.

\subsection{Dataset Processing}~\label{app:dataset_processing}

Our processing protocol followed these principles:

\begin{enumerate}
    \item \textbf{Split Handling:} If a dataset contains original train/dev/test splits, we adhered to them even if some splits were missing. For datasets with only a train split, we implemented an 80/10/10 split for consistency.
    \item \textbf{Filtering Criteria:} To maintain quality, we filtered out data instances with more than 15 extracted event labels, especially in MAVEN-Arg, where instances with excessive labels could skew results.
    \item \textbf{Query Length Diversity:} We ensured the datasets included varied lengths of queries, from single sentences to longer documents, enriching the task complexity and addressing different real-world scenarios.
\end{enumerate}

\subsection{Greedy Maximum Independent Set}~\label{app:GMIS}

Algorithm~\ref{alg:greedy_max_indep_set} shows the details of Greedy Maximum Independent Set algorithm for identifying the largest possible subset of diverse schemas.

\begin{algorithm}[H]
\caption{Greedy Maximum Independent Set}
\label{alg:greedy_max_indep_set}
\begin{algorithmic}[1]
\REQUIRE Adjacency list \( \text{adj\_list} \) representing schema similarities
\ENSURE Maximum independent set of schemas

\STATE \( \text{remaining} \gets \{ \text{all schema indices in } \text{adj\_list} \} \)
\STATE \( \text{independent\_set} \gets \{\} \)
\STATE \( \text{degrees} \gets \{ i: \text{len}(\text{adj\_list}[i]) \mid i \in \text{adj\_list} \} \)

\WHILE{remaining is not empty}
    \STATE \( \text{node} \gets \text{argmin}_{i \in \text{remaining}}(\text{degrees}[i]) \)
    \STATE \( \text{independent\_set} \gets \text{independent\_set} \cup \{ \text{node} \} \)
    \STATE \( \text{remaining} \gets \text{remaining} \setminus \{ \text{node} \} \)
    \STATE \( \text{remaining} \gets \text{remaining} \setminus \text{adj\_list}[\text{node}] \)

    \FOR{each neighbor \( n \in \text{adj\_list}[\text{node}] \)}
        \STATE \( \text{degrees} \gets \text{degrees} \setminus \{ n \} \)
        \FOR{each neighbor \( m \in \text{adj\_list}[n] \)}
            \IF{m in remaining}
                \STATE \( \text{degrees}[m] \gets \text{degrees}[m] - 1 \)
            \ENDIF
        \ENDFOR
    \ENDFOR
    \STATE \( \text{degrees} \gets \text{degrees} \setminus \{ \text{node} \} \)
\ENDWHILE

\RETURN \( \text{independent\_set} \)
\end{algorithmic}
\end{algorithm}

\subsection{Dataset Statistics}~\label{app:dataset_statistics}
Table~\ref{tab:dataset_statistics_1} shows the statistics of the collected datasets, such as CrudeOilNews, GENEVA, MAVEN-Arg, DocEE-en, DocEE-zh, IEPILE-en, and IEPILE-zh, including language, domain, split distributions, and maximum number of labels per sample.
Table~\ref{tab:dataset_statistics_2} shows the statistics of the MD-SEE dataset, including source, language, domain, split distributions, and maximum number of labels per sample.

\begin{table*}[tbp]
    \footnotesize 
    \centering 
    \addtolength{\tabcolsep}{-2pt}
    \begin{tabular}{c|c|c|c|ccccc}
        \toprule
        Dataset& Source       & Language & Domain        & \#Schemas & \#Train & \#Dev & \#Test & \#Max\_Labels \\ \midrule
        CrudeOilNews& -  & en       & Oil News          & 18        & 1489    & -   & 265    & 10         \\ 
        GENEVA&  -      & en       & General       & 115       & 1922    & 778  & 931   & 10         \\ 
        MAVEN-Arg&  -    & en       & General       & 162       & 2913    & -     & 710    & 15         \\
        DocEE-en& DocEE         & en       & General       & 59        & 21966   & 2748  & 2771   & 1          \\ 
        DocEE-zh& DocEE        & zh       & General       & 58       & 29383   & 3672  & 3674   & 1          \\
        \midrule
        \multirow{4}{*}{IEPILE-en} & CASIE$^\dagger$         & en       & Cybersecurity & 5         & 3732    & 777   & 1492   & 1          \\
         & PHEE$^\dagger$          & en       & Biomedical    & 2         & 2897   & 960  & 968   & 1          \\ 
         & RAMS$^\dagger$          & en       & News          & 106       & -       & -     & 887    & 1          \\ 
         & WikiEvents$^\dagger$    & en       & Wikipedia     & 31        & -       & -     & 249    & 5          \\
        \midrule
        \multirow{4}{*}{IEPILE-zh}  & DuEE-fin$^\dagger$      & zh       & Finance       & 13        & 7015    & -   & 1171   & 14         \\ 
        & DuEE1.0$^\dagger$       & zh       & News          & 65        & 11908   & -   & 1492   & 15         \\
        & FewFC$^\dagger$         & zh       & Finance       & 5         & -       & -     & 2879   & 5          \\ 
        & ccf\_law$^\dagger$      & zh       & Legal         & 9        & -       & -     & 971   & 10         \\ 
        \bottomrule
    \end{tabular}
    \caption{Statistics of the collected datasets including language, domain, split distributions, and maximum number of labels per sample. Datasets marked with $\dagger$ were collected from IEPILE and further processed in this work.}
    \label{tab:dataset_statistics_1}
\end{table*}

\begin{table*}[tbp]
    \footnotesize 
    \centering 
    \addtolength{\tabcolsep}{-2pt}
    \begin{tabular}{c|c|c|ccccc}
        \toprule
        Source       & Language & Domain        & \#Schemas & \#Train & \#Dev & \#Test & \#Max\_Labels \\ \midrule
        DocEE         & en/zh       & General       & 52        & 2000    & 200   & 1000   & 1          \\ 
        DuEE1.0       & zh/en       & News          & 53        & 2000    & -   & 1000    & 8          \\ 
        CrudeOilNews  & en       & Oil News          & 11        & 949    & -   & 198    & 6          \\
        GENEVA        & en       & General       & 61        & 1278    & 200   & 600    & 7          \\ 
        MAVEN-Arg     & en       & General       & 30        & 590     & -     & 150    & 8          \\ 
        CASIE         & en       & Cybersecurity & 4         & 2000    & 200   & 1000    & 1          \\ 
        PHEE          & en       & Biomedical    & 1         & 2000     & 200   & 867    & 1          \\ 
        RAMS          & en       & News          & 52        & -       & -     & 286    & 1          \\ 
        WikiEvents    & en       & Wikipedia     & 18        & -       & -     & 66     & 2          \\
        DuEE-fin      & zh       & Finance       & 7         & 2000    & -   & 558    & 11         \\ 
        FewFC         & zh       & Finance       & 5         & -       & -     & 1000   & 5          \\
        ccf\_law      & zh       & Legal         & 6         & -       & -     & 961    & 9          \\ 
        \midrule
        \textbf{Total}         & -                 & -                      & 300                & 12817            & 800           & 7686            & 11                  \\ 
        \bottomrule
    \end{tabular}
    \caption{Statistics of the MD-SEE dataset, including source, language, domain, split distributions, and maximum number of labels per sample. Mixed languages (en/zh or zh/en) indicate the language of queries and schemas respectively, representing subsets for cross-lingual extraction.}
    \label{tab:dataset_statistics_2}
\end{table*}

\section{End-to-End Evaluation Experimental Results}~\label{app:end2end evaluation}

The following tables show the end-to-end results of CrudeOilNews (Table~\ref{tab:End to end results of CrudeOilNews}), DocEE-en (Table~\ref{tab:End to end results of DocEE-en}), DocEE-zh (Table~\ref{tab:End to end results of DocEE-zh}), GENEVA (Table~\ref{tab:End to end results of GENEVA}), IEPILE-en (Table~\ref{tab:End to end results of IEPILE_EE_en}), IEPILE-zh (Table~\ref{tab:End to end results of IEPILE_EE_zh}), and MAVEN-Arg (Table~\ref{tab:End to end results of MAVEN-Arg}).

\begin{table*}[ht]
\centering
\addtolength{\tabcolsep}{-4.5pt}
\resizebox{0.88\textwidth}{!}{
\begin{tabular}{lcccccccc}
    \toprule
    & Phi-3.5-mini & Llama-3.2-3B & Llama-3.1-8B & Mistral-7B-v0.3 & Qwen2.5-7B & Qwen2.5-14B & YAYI-UIE & GPT-4-turbo \\
    \midrule
    \multicolumn{9}{c}{Raw} \\
    \midrule
    BM25 & 0.08 & 0.12 & 0.15 & 0.13 & 0.15 & 0.10 & 0.12 & 0.17 \\
    BGE-M3 & 0.08 & 0.12 & 0.15 & 0.13 & 0.15 & 0.10 & 0.12 & 0.17 \\
    E5-LV2 & 0.08 & 0.12 & 0.15 & 0.13 & 0.15 & 0.10 & 0.12 & 0.17 \\
    GTE-L & 0.07 & 0.12 & 0.14 & 0.12 & 0.14 & 0.10 & 0.12 & 0.17 \\
    LLM-E & 0.07 & 0.11 & 0.13 & 0.11 & 0.13 & 0.09 & 0.11 & 0.15 \\
    BGE-RB & 0.07 & 0.11 & 0.13 & 0.11 & 0.13 & 0.09 & 0.11 & 0.15 \\
    BGE-RL & 0.06 & 0.10 & 0.12 & 0.10 & 0.12 & 0.08 & 0.10 & 0.14 \\
    \midrule
    \multicolumn{9}{c}{Paraph.} \\
    \midrule
    BM25 & 0.07 & 0.11 & 0.13 & 0.11 & 0.13 & 0.09 & 0.11 & 0.15 \\
    BGE-M3 & 0.07 & 0.11 & 0.14 & 0.12 & 0.14 & 0.10 & 0.11 & 0.16 \\
    E5-LV2 & 0.08 & 0.13 & 0.16 & 0.13 & 0.15 & 0.11 & 0.13 & 0.18 \\
    GTE-L & 0.08 & 0.13 & 0.16 & 0.14 & 0.15 & 0.11 & 0.13 & 0.18 \\
    LLM-E & 0.07 & 0.11 & 0.14 & 0.12 & 0.13 & 0.09 & 0.11 & 0.16 \\
    BGE-RB & 0.07 & 0.11 & 0.14 & 0.12 & 0.14 & 0.10 & 0.11 & 0.16 \\
    BGE-RL & 0.07 & 0.11 & 0.13 & 0.11 & 0.13 & 0.09 & 0.11 & 0.15 \\
    \bottomrule
\end{tabular}}
    \caption{End to end results of CrudeOilNews.}
    \label{tab:End to end results of CrudeOilNews}
\end{table*}

\begin{table*}[ht]
\centering
\addtolength{\tabcolsep}{-4.5pt}
\resizebox{0.88\textwidth}{!}{
\begin{tabular}{lcccccccc}
    \toprule
    & Phi-3.5-mini & Llama-3.2-3B & Llama-3.1-8B & Mistral-7B-v0.3 & Qwen2.5-7B & Qwen2.5-14B & YAYI-UIE & GPT-4-turbo \\
    \midrule
    \multicolumn{9}{c}{Raw} \\
    \midrule
    BM25 & 0.09 & 0.10 & 0.10 & 0.09 & 0.09 & 0.10 & 0.08 & 0.12 \\
    BGE-M3 & 0.36 & 0.40 & 0.41 & 0.38 & 0.38 & 0.41 & 0.31 & 0.47 \\
    E5-LV2 & 0.36 & 0.40 & 0.40 & 0.38 & 0.38 & 0.41 & 0.31 & 0.47 \\
    GTE-L & 0.41 & 0.45 & 0.46 & 0.43 & 0.43 & 0.47 & 0.35 & 0.53 \\
    LLM-E & 0.38 & 0.41 & 0.42 & 0.39 & 0.39 & 0.43 & 0.32 & 0.49 \\
    BGE-RB & 0.19 & 0.20 & 0.21 & 0.19 & 0.19 & 0.21 & 0.16 & 0.24 \\
    BGE-RL & 0.25 & 0.27 & 0.27 & 0.26 & 0.26 & 0.28 & 0.21 & 0.32 \\
    \midrule
    \multicolumn{9}{c}{Paraph.} \\
    \midrule
    BM25 & 0.29 & 0.31 & 0.32 & 0.30 & 0.30 & 0.33 & 0.25 & 0.37 \\
    BGE-M3 & 0.39 & 0.43 & 0.44 & 0.41 & 0.41 & 0.45 & 0.34 & 0.51 \\
    E5-LV2 & 0.37 & 0.41 & 0.42 & 0.39 & 0.39 & 0.42 & 0.32 & 0.49 \\
    GTE-L & 0.41 & 0.45 & 0.45 & 0.43 & 0.43 & 0.46 & 0.35 & 0.53 \\
    LLM-E & 0.38 & 0.42 & 0.43 & 0.40 & 0.40 & 0.44 & 0.33 & 0.50 \\
    BGE-RB & 0.21 & 0.23 & 0.23 & 0.22 & 0.22 & 0.24 & 0.18 & 0.27 \\
    BGE-RL & 0.33 & 0.36 & 0.37 & 0.34 & 0.34 & 0.37 & 0.28 & 0.43 \\
    \bottomrule
\end{tabular}}
    \caption{End to end results of DocEE-en.}
    \label{tab:End to end results of DocEE-en}
\end{table*}

\begin{table*}[ht]
\centering
\addtolength{\tabcolsep}{-4.5pt}
\resizebox{0.88\textwidth}{!}{
\begin{tabular}{lcccccccc}
    \toprule
    & Phi-3.5-mini & Llama-3.2-3B & Llama-3.1-8B & Mistral-7B-v0.3 & Qwen2.5-7B & Qwen2.5-14B & YAYI-UIE & GPT-4-turbo \\
    \midrule
    \multicolumn{9}{c}{Raw} \\
    \midrule
    BM25 & 0.25 & 0.33 & 0.33 & 0.31 & 0.22 & 0.34 & 0.16 & 0.31 \\
    BGE-M3 & 0.34 & 0.44 & 0.45 & 0.42 & 0.30 & 0.45 & 0.22 & 0.44 \\
    E5-LV2 & 0.19 & 0.24 & 0.25 & 0.23 & 0.17 & 0.25 & 0.12 & 0.23 \\
    GTE-L & 0.22 & 0.28 & 0.29 & 0.27 & 0.19 & 0.29 & 0.14 & 0.26 \\
    LLM-E & 0.16 & 0.20 & 0.21 & 0.19 & 0.14 & 0.21 & 0.10 & 0.19 \\
    BGE-RB & 0.22 & 0.29 & 0.29 & 0.27 & 0.20 & 0.30 & 0.14 & 0.28 \\
    BGE-RL & 0.31 & 0.40 & 0.40 & 0.38 & 0.27 & 0.41 & 0.21 & 0.37 \\
    \midrule
    \multicolumn{9}{c}{Paraph.} \\
    \midrule
    BM25 & 0.36 & 0.46 & 0.47 & 0.44 & 0.32 & 0.48 & 0.23 & 0.43 \\
    BGE-M3 & 0.43 & 0.56 & 0.57 & 0.53 & 0.38 & 0.58 & 0.28 & 0.53 \\
    E5-LV2 & 0.24 & 0.30 & 0.31 & 0.29 & 0.21 & 0.31 & 0.15 & 0.28 \\
    GTE-L & 0.23 & 0.28 & 0.29 & 0.27 & 0.21 & 0.29 & 0.14 & 0.27 \\
    LLM-E & 0.15 & 0.19 & 0.20 & 0.18 & 0.14 & 0.20 & 0.10 & 0.18 \\
    BGE-RB & 0.20 & 0.26 & 0.27 & 0.25 & 0.18 & 0.27 & 0.13 & 0.25 \\
    BGE-RL & 0.26 & 0.35 & 0.36 & 0.33 & 0.28 & 0.38 & 0.21 & 0.43 \\
    \bottomrule
\end{tabular}}
    \caption{End to end results of DocEE-zh.}
    \label{tab:End to end results of DocEE-zh}
\end{table*}

\begin{table*}[ht]
\centering
\addtolength{\tabcolsep}{-4.5pt}
\resizebox{0.88\textwidth}{!}{
\begin{tabular}{lcccccccc}
    \toprule
    & Phi-3.5-mini & Llama-3.2-3B & Llama-3.1-8B & Mistral-7B-v0.3 & Qwen2.5-7B & Qwen2.5-14B & YAYI-UIE & GPT-4-turbo \\
    \midrule
    \multicolumn{9}{c}{Raw} \\
    \midrule
    BM25 & 0.03 & 0.05 & 0.06 & 0.06 & 0.07 & 0.05 & 0.05 & 0.07 \\
    BGE-M3 & 0.13 & 0.20 & 0.25 & 0.24 & 0.26 & 0.21 & 0.18 & 0.29 \\
    E5-LV2 & 0.13 & 0.20 & 0.25 & 0.24 & 0.26 & 0.21 & 0.18 & 0.29 \\
    GTE-L & 0.13 & 0.19 & 0.23 & 0.22 & 0.24 & 0.20 & 0.15 & 0.27 \\
    LLM-E & 0.12 & 0.19 & 0.23 & 0.23 & 0.24 & 0.20 & 0.15 & 0.27 \\
    BGE-RB & 0.10 & 0.16 & 0.19 & 0.18 & 0.20 & 0.15 & 0.13 & 0.22 \\
    BGE-RL & 0.13 & 0.19 & 0.24 & 0.23 & 0.25 & 0.20 & 0.15 & 0.27 \\
    \midrule
    \multicolumn{9}{c}{Paraph.} \\
    \midrule
    BM25 & 0.13 & 0.19 & 0.24 & 0.23 & 0.25 & 0.20 & 0.18 & 0.27 \\
    BGE-M3 & 0.16 & 0.24 & 0.29 & 0.29 & 0.30 & 0.25 & 0.22 & 0.34 \\
    E5-LV2 & 0.16 & 0.24 & 0.29 & 0.29 & 0.31 & 0.25 & 0.22 & 0.34 \\
    GTE-L & 0.15 & 0.22 & 0.28 & 0.27 & 0.29 & 0.24 & 0.20 & 0.32 \\
    LLM-E & 0.16 & 0.24 & 0.29 & 0.29 & 0.31 & 0.25 & 0.22 & 0.34 \\
    BGE-RB & 0.14 & 0.22 & 0.27 & 0.26 & 0.28 & 0.23 & 0.20 & 0.32 \\
    BGE-RL & 0.13 & 0.19 & 0.24 & 0.24 & 0.25 & 0.20 & 0.18 & 0.27 \\
    \bottomrule
\end{tabular}}
    \caption{End to end results of GENEVA.}
    \label{tab:End to end results of GENEVA}
\end{table*}

\begin{table*}[ht]
\centering
\addtolength{\tabcolsep}{-4.5pt}
\resizebox{0.88\textwidth}{!}{
\begin{tabular}{lcccccccc}
    \toprule
    & Phi-3.5-mini & Llama-3.2-3B & Llama-3.1-8B & Mistral-7B-v0.3 & Qwen2.5-7B & Qwen2.5-14B & YAYI-UIE & GPT-4-turbo \\
    \midrule
    \multicolumn{9}{c}{Raw} \\
    \midrule
    BM25 & 0.20 & 0.27 & 0.33 & 0.30 & 0.33 & 0.32 & 0.25 & 0.38 \\
    BGE-M3 & 0.25 & 0.34 & 0.42 & 0.38 & 0.42 & 0.41 & 0.31 & 0.47 \\
    E5-LV2 & 0.27 & 0.37 & 0.46 & 0.41 & 0.46 & 0.44 & 0.34 & 0.51 \\
    GTE-L & 0.28 & 0.37 & 0.46 & 0.41 & 0.46 & 0.44 & 0.34 & 0.52 \\
    LLM-E & 0.27 & 0.36 & 0.45 & 0.40 & 0.45 & 0.43 & 0.33 & 0.50 \\
    BGE-RB & 0.22 & 0.29 & 0.36 & 0.32 & 0.36 & 0.35 & 0.27 & 0.41 \\
    BGE-RL & 0.26 & 0.34 & 0.43 & 0.38 & 0.43 & 0.41 & 0.32 & 0.48 \\
    \midrule
    \multicolumn{9}{c}{Paraph.} \\
    \midrule
    BM25 & 0.26 & 0.34 & 0.43 & 0.38 & 0.43 & 0.41 & 0.32 & 0.48 \\
    BGE-M3 & 0.29 & 0.39 & 0.49 & 0.43 & 0.49 & 0.47 & 0.36 & 0.55 \\
    E5-LV2 & 0.30 & 0.40 & 0.50 & 0.44 & 0.50 & 0.48 & 0.37 & 0.56 \\
    GTE-L & 0.29 & 0.39 & 0.49 & 0.43 & 0.49 & 0.47 & 0.36 & 0.55 \\
    LLM-E & 0.29 & 0.39 & 0.49 & 0.43 & 0.49 & 0.47 & 0.36 & 0.55 \\
    BGE-RB & 0.27 & 0.37 & 0.46 & 0.41 & 0.46 & 0.43 & 0.33 & 0.52 \\
    BGE-RL & 0.27 & 0.36 & 0.44 & 0.40 & 0.45 & 0.43 & 0.33 & 0.50 \\
    \bottomrule
\end{tabular}}
    \caption{End to end results of IEPILE-en.}
    \label{tab:End to end results of IEPILE_EE_en}
\end{table*}

\begin{table*}[ht]
\centering
\addtolength{\tabcolsep}{-4.5pt}
\resizebox{0.88\textwidth}{!}{
\begin{tabular}{lcccccccc}
    \toprule
    & Phi-3.5-mini & Llama-3.2-3B & Llama-3.1-8B & Mistral-7B-v0.3 & Qwen2.5-7B & Qwen2.5-14B & YAYI-UIE & GPT-4-turbo \\
    \midrule
    \multicolumn{9}{c}{Raw} \\
    \midrule
    BM25 & 0.45 & 0.53 & 0.53 & 0.53 & 0.55 & 0.58 & 0.25 & 0.57 \\
    BGE-M3 & 0.50 & 0.59 & 0.59 & 0.59 & 0.61 & 0.65 & 0.28 & 0.64 \\
    E5-LV2 & 0.28 & 0.33 & 0.33 & 0.33 & 0.34 & 0.36 & 0.16 & 0.36 \\
    GTE-L & 0.20 & 0.24 & 0.24 & 0.24 & 0.25 & 0.27 & 0.12 & 0.26 \\
    LLM-E & 0.16 & 0.19 & 0.19 & 0.19 & 0.20 & 0.21 & 0.09 & 0.19 \\
    BGE-RB & 0.54 & 0.63 & 0.63 & 0.63 & 0.65 & 0.70 & 0.30 & 0.69 \\
    BGE-RL & 0.54 & 0.63 & 0.63 & 0.63 & 0.65 & 0.70 & 0.30 & 0.69 \\
    \midrule
    \multicolumn{9}{c}{Paraph.} \\
    \midrule
    BM25 & 0.49 & 0.58 & 0.58 & 0.58 & 0.59 & 0.63 & 0.28 & 0.62 \\
    BGE-M3 & 0.55 & 0.65 & 0.65 & 0.65 & 0.67 & 0.71 & 0.31 & 0.70 \\
    E5-LV2 & 0.18 & 0.21 & 0.21 & 0.21 & 0.22 & 0.23 & 0.10 & 0.23 \\
    GTE-L & 0.26 & 0.31 & 0.31 & 0.31 & 0.32 & 0.34 & 0.14 & 0.34 \\
    LLM-E & 0.32 & 0.37 & 0.37 & 0.37 & 0.39 & 0.41 & 0.18 & 0.40 \\
    BGE-RB & 0.53 & 0.62 & 0.62 & 0.62 & 0.64 & 0.68 & 0.29 & 0.67 \\
    BGE-RL & 0.54 & 0.64 & 0.64 & 0.64 & 0.67 & 0.70 & 0.30 & 0.69 \\
    \bottomrule
\end{tabular}}
    \caption{End to end results of IEPILE-zh.}
    \label{tab:End to end results of IEPILE_EE_zh}
\end{table*}

\begin{table*}[ht]
\centering
\addtolength{\tabcolsep}{-4.5pt}
\resizebox{0.88\textwidth}{!}{
\begin{tabular}{lcccccccc}
    \toprule
    & Phi-3.5-mini & Llama-3.2-3B & Llama-3.1-8B & Mistral-7B-v0.3 & Qwen2.5-7B & Qwen2.5-14B & YAYI-UIE & GPT-4-turbo \\
    \midrule
    \multicolumn{9}{c}{Raw} \\
    \midrule
    BM25 & 0.03 & 0.04 & 0.05 & 0.04 & 0.04 & 0.04 & 0.03 & 0.05 \\
    BGE-M3 & 0.04 & 0.05 & 0.07 & 0.06 & 0.06 & 0.05 & 0.04 & 0.08 \\
    E5-LV2 & 0.06 & 0.07 & 0.10 & 0.09 & 0.09 & 0.08 & 0.06 & 0.11 \\
    GTE-L & 0.06 & 0.07 & 0.10 & 0.09 & 0.09 & 0.08 & 0.06 & 0.11 \\
    LLM-E & 0.05 & 0.06 & 0.09 & 0.08 & 0.08 & 0.07 & 0.06 & 0.09 \\
    BGE-RB & 0.03 & 0.04 & 0.05 & 0.05 & 0.05 & 0.04 & 0.03 & 0.06 \\
    BGE-RL & 0.04 & 0.05 & 0.07 & 0.06 & 0.07 & 0.06 & 0.05 & 0.08 \\
    \midrule
    \multicolumn{9}{c}{Paraph.} \\
    \midrule
    BM25 & 0.06 & 0.08 & 0.11 & 0.10 & 0.10 & 0.09 & 0.07 & 0.12 \\
    BGE-M3 & 0.10 & 0.12 & 0.17 & 0.16 & 0.16 & 0.13 & 0.11 & 0.18 \\
    E5-LV2 & 0.09 & 0.12 & 0.16 & 0.15 & 0.15 & 0.12 & 0.10 & 0.17 \\
    GTE-L & 0.09 & 0.12 & 0.16 & 0.15 & 0.15 & 0.13 & 0.10 & 0.18 \\
    LLM-E & 0.09 & 0.12 & 0.16 & 0.15 & 0.15 & 0.12 & 0.10 & 0.17 \\
    BGE-RB & 0.06 & 0.08 & 0.11 & 0.10 & 0.10 & 0.09 & 0.07 & 0.13 \\
    BGE-RL & 0.06 & 0.08 & 0.10 & 0.10 & 0.10 & 0.08 & 0.07 & 0.11 \\
    \bottomrule
\end{tabular}}
    \caption{End to end results of MAVEN-Arg.}
    \label{tab:End to end results of MAVEN-Arg}
\end{table*}

\end{document}